# ENWalk: Learning Network Features for Spam Detection in Twitter

Santosh K C[1], Suman Kalyan Maity[2], Arjun Mukherjee[1]

[1] University of Houston, [2] IIT Kharagpur
skc@uh.edu, sumankalyan.maity@cse.iitkgp.ernet.in, arjun@uh.edu

**Abstract.** Social medias are increasing their influence with the vast public information leading to their active use for marketing by the companies and organizations. Such marketing promotions are difficult to identify unlike the traditional medias like TV and newspaper. So, it is very much important to identify the promoters in the social media. Although, there are active ongoing researches, existing approaches are far from solving the problem. To identify such imposters, it is very much important to understand their strategies of social circle creation and dynamics of content posting. Are there any specific spammer types? How successful are each types? We analyze these questions in the light of social relationships in Twitter. Our analyses discover two types of spammers and their relationships with the dynamics of content posts. Our results discover novel dynamics of spamming which are intuitive and arguable. We propose *ENWalk*, a framework to detect the spammers by learning the feature representations of the users in the social media. We learn the feature representations using the random walks biased on the spam dynamics. Experimental results on large-scale twitter network and the corresponding tweets show the effectiveness of our approach that outperforms the existing approaches.

## 1 Introduction

Social medias are increasing their influence tremendously. Twitter is one of the popular platforms where people post information in the form of tweets and share the tweets. Twitter is available from wide range of web-enabled services to all the people. So, the real time reflection of a society can be viewed in twitter. Celebrities, governments, politicians, businesses are active in twitter to provide their updates and to listen to the views of the people. Thus, the bidirectional flow of information is high. The openness of the online platforms and reliance on users facilitates the spammers to easily penetrate the platform and overwhelm the users with malicious intent and content. This work attempts to detect the spammers in social network using a case study of twitter.

Spammers in social networks constantly adapt to avoid the detection. Moreover, they follow reflexive reciprocity [6, 17] (users following back when they are followed by someone to show courtesy) to establish social influence and act normal. So, it is becoming difficult for traditional spam detection methods to detect the spammers. Such spammers have widespread impacts. There are several reports of army of fake Twitter accounts[1] being used to troll[2] and promote political agendas[3]. Even US President Donald Trump has been accused of fake followers[4].

In this paper, we present ENWalk, a framework that uses the content information to bias a random walk of the network and obtain the latent feature embedding of the nodes in the network. ENWalk generates the biased random walks and uses them to maximize the likelihood of obtaining similar nodes in the neighborhood of the network. We study the twitter content dynamics that could be important to bias those random walks. We found that there are two types of spammers: *follow-flood* and *vigilant*. We found that success rate, activity window, fraudulence and mentioning behaviors can be used to compare the equivalence of users in the twitter. We calculate the network equivalence using these four behavioral features between pairs of nodes and try to bias the random walks with interaction proximity of the pair of nodes. Experimental results on 17 million user network from twitter show that the combination of behavioral features with the underlying network structure significantly outperform the existing state-of-the-art approaches for deception detection.

## 2   Related Work

There have been several works on spam detection in general, especially review spam [7], and opinion spam. However, in Twitter there are limited attempts. One of the earliest works was done by Benevenuto et al. [1]. They manually labeled and trained a traditional classifier using the features extracted from user contents and behaviors. Lee et al. leveraged profile-based features and deployed social honeypots to detect new social spammers [9]. Stringhini et al. also studied spam detection using honey profiles [14]. Ghosh et al. studied the problem of link farming in Twitter [4] and introduced a ranking methodology to penalize the link farmers. Abuse of online social networks was studied in [16]. Campaign spams was studied on [3, 10, 19].

Skip-gram model [12] has been popular to learn the features from a large corpus of data. It inspired to establish an analogy for networks by representing a network as a "document". Similar to document being an ordered sequence of words, we can create an ordered sequence of nodes from a network using sampling techniques. DeepWalk [13] learns d-dimensional feature representations by simulating uniform random walks. LINE [15] learns the d-dimensional features into two phases: d/2 BFS-style simulations and another d/2 2-hop distant nodes. Node2vec [5] creates the ordered sequence simulating the BFS and DFS approaches. All these feature learning approaches don't use the data associated with node which are important to learn the behaviors of the nodes.

---

[1] http://theatln.tc/2m8g3eA
[2] http://bzfd.it/2m8rlja
[3] http://bit.ly/2kJiMKu
[4] http://bit.ly/1ViorHd, http://53eig.ht/2kzrhfL

## 3 Dataset

For this work, we use the Twitter dataset used in [18]. It contains 17 million users having 467 million Twitter posts covering a seven month period from June 1 2009 to December 31 2009. To extract the network graph for those 17 million users, we extracted the follower-following topology of Twitter from [8] which contains all the entire twitter user profiles and their social relationships till July 2009. We pruned the users so that they have social relationship in [8] and tweets in [18] and are left with 4,405,698 users. Twitter suspends the accounts involved in the malicious activity (*https://support.twitter.com/articles/18311*). To obtain the suspend status of accounts, we re-crawled the profile pages of all the 17 million users. This yielded a total of 100,758 accounts that had been suspended (the profile page redirects to the page *https://twitter.com/account/suspended*). We use this suspension signal as the primary signal for evaluating our models as the primary reason for account suspension is the involvement in the spam activity. However, there might be other reasons like inactivity. So, to ensure the suspended accounts are spammers, we further checked for malicious activities for those users. For this, we examined various URLs from the account's timeline and checked them against a list of blacklisted URLs. We use three blacklists: Google Safebrowsing (*http://code.google.com/apis/safebrowsing/*), URIBL (*http://uribl.com/*) and Joewein (*http://www.joewein.net/*). We found that 75% of suspended accounts posted at least one shortened URL blacklisted. We also looked for duplicate tweets enforced for promotion. After applying these additional criteria, our final data comprised of 86,652 spammers and 4,319,046 non-spammers, which was used for evaluating our model.

## 4 Spam Analysis

Characterizing the dominant spammer types is important as it is the first step in understanding the dynamics of spamming. We studied the follower-following network creation strategies of the spammers. We found that there are two main types of spamming based on the follow-following strategies: (1) *follow-flood* spammers and (2) *vigilant* spammers. So, the question arises why some spammers are more successful? In this section, we study the behavioral aspects of tweet dynamics of spammers. We later leverage them in model building.

### 4.1 Spammer Type

To analyze the strategies of follower-following, we calculated the number of followers (users that are following the current user) and the number of followings (users that the current user is following) for each spammer. Figure 1 shows the plot in log scale count. It shows that the follower and following count differ for each spammers. The users with more followers than followings tend to be more successful as they have been able to "earn" a lot of users who are following them. So, we define success rate as:

$$sr_u = \frac{\text{\# of followers of } u}{\text{\# of followings of } u} \quad (1)$$

Based on the network expansion success rate, we find that there are two dominant spamming strategies:

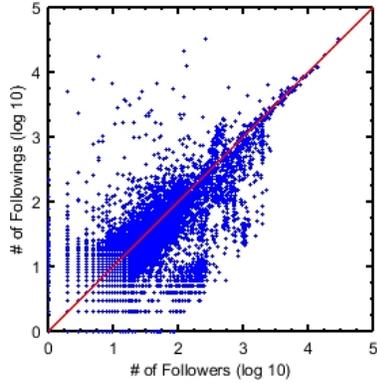

**Figure 1. Follower-Following Count of Spammers. Each Blue dot represents a spammer and the red line is the plot of y=x line.**

- *Follow-flood* Spammers: These are less successful spammers who just flood the network with friendship initiation so as to get followers who they can influence. We categorize the spammers with success rate ($sr_u$) less than 1 in this type.
- *Vigilant* Spammers: These are successful spammers who take a cautious approach of friendship creation and content posting. Spammers with success rate ($sr_u$) greater or equal to 1 are categorized as *vigilant*.

To learn the dynamics of each spammer type, we further analyzed the success rate of spammers with other behavioral aspects – activity window, usage of promotion words or blacklist words and hashtag mentioning.

### 4.2 Activity Window

We compute the activity window as the number of days a user is active in the twitter network. Since, we don't have the exact time when a user was suspended, we approximate the time of suspension as the date of the last tweet tweeted by the user. We found that the average activity window of a *vigilant* spammer is 138 days with a standard deviation of 19 days compared to the average of 35 days and standard deviation of 12 days for *follow-flood* spammers. Although, the basic strategy of any spammer is to inject itself into the network and emit the spam contents, the success rate also depends how long it can remain undetected in the network. So, *vigilant* spammers have a higher success rate.

### 4.3 Fraudulence

One of the primary reason to spam is to inject constant fraudulence information. So, we analyzed the fraudulence behavior of the two types of spammers. We labeled the tweets containing promotional, adult words or the blacklisted urls as fraud tweets. So, we compute fraudulence as:

$$fr_u = \frac{\# \text{ of fraud tweets of } u}{\text{total \# of tweets of } u} \quad (2)$$

We found that the average fraudulence of *vigilant* spammers is 0.34 compared to 0.86 of *follow-flood* spammers. So, the follow-flood spammers are more involved in spam.

### 4.4 Mentioning Celebrities and Popular Hashtags

Mentioning the popular celebrities or hashtags empowers a tweet. So, one of the common strategies of spammers is to include the popular ones in their tweets. We studied mentioning phenomenon and found that *vigilant* spammers mention half the celebrities per tweets compared to the *follow-flood* spammers.

## 5 Learning Latent Features for Spam Detection

Having characterized the dynamics of spamming in Twitter, can we improve spam detection beyond the existing state-of-the-art approaches? To answer this we used our Twitter data to setup a latent feature learning problem in networks. Our analysis is general and can be used to any social network.

### 5.1 Overview

As discussed in the previous section, the dynamics of Twitter are interesting and can be leveraged to catch the spammers. So, we use the spam dynamics to formulate the latent feature learning in social networks. Let $G = (V, E, X)$ be a given network with vertices, edges and the social network data of users in the social network. We aim to learn a mapping function $f : V \to \mathbb{R}^d$ from nodes to a d-dimensional feature representations which can be used for prediction. The parameter $d$ specifies the number of dimensions of the latent features such that the size of $f$ is $|V| \times d$.

We present a novel sampling strategy that samples nodes in network exploiting the spam dynamics such that the *equivalent neighborhood* $EN(u) \subset V$ contains the node having similar tweeting behaviors with the node $u$. We generate $EN(u)$ for each nodes in the network and predict which nodes are the members of $u$'s equivalent neighbors based on the learnt latent features $f$. The basic rationale is that we wish to learn latent feature representations for nodes that respect equivalent neighborhoods (which are based on the spamming dynamics) so that classification/ranking using the learned representation yields results that leverage the spamming dynamics.

### 5.2 The Optimization Problem

As our goal is to learn the latent features $f$ that best describe the equivalent neighborhood $EN(u)$ of node $u$, we define the optimization problem as follows:

$$\max_f \sum_{u \in V} \log Pr\big(EN(u)\big|f(u)\big) \qquad (3)$$

To solve the optimization problem, we extend the SkipGram architecture [5, 13, 15] which approximates the conditional probability using an independence assumption that the likelihood of observing an equivalent neighborhood node is independent of observing any other equivalent neighborhood given the latent features of the source node.

$$Pr\big(EN(u)\big|f(u)\big) = \prod_{v \in EN(u)} \Pr(v|f(u)) \qquad (4)$$

Since, the source node and the equivalent neighborhood node have symmetric equivalence, the conditional likelihood can be modeled as softmax unit parameterized by a dot product of their features.

$$Pr\big(v|f(u)\big) = \frac{exp(\,f(v).\,f(u))}{\sum_{t \in V} exp(f(t).\,f(u))} \qquad (5)$$

The optimization problem now becomes:

$$\max_{f} \sum_{u \in V} \left[ -\log Z_u + \sum_{t \in EN(u)} f(t) \cdot f(u) \right] \quad (6)$$

For large networks, the partition function $Z_u = \sum_{t \in V} \exp(f(t) \cdot f(u))$ is expensive to compute. So, we use negative sampling [12] to approximate it. We use stochastic gradient descent over the model parameters defining the features $f$. Feature learning methods based on Skip-gram architecture are developed for natural language [11]. Since natural language texts are linear, the notion of a neighborhood can be naturally defined using a sliding window over consecutive words in sentences. Networks are not linear, and thus a richer notion of a neighborhood is needed. To mitigate this problem, we use multiple biased random walks each one in principle exploring a different neighborhood [5].

### 5.3 Equivalent Neighborhood Generation

The analyses of spam dynamics leads to an important inference that the nodes are similar if they have similar spam dynamics. So, we want to exploit those dynamics to generate the equivalent neighborhood $EN(u)$ for the node $u$. Nodes in a network are equivalent if they share similar behaviors. We use the random walk procedure which can be biased to generate the equivalent neighborhood.

We bias the random walks based on the four dynamics: common time of activity ($ct_{tv}$), success rate difference ($sr_{tv}$), fraudulence commonalities ($fr_{tv}$) and common mentioning in tweets ($me_{tv}$). We calculate each dynamics as follows:

$$ct_{tv} = \frac{\text{\# of days with common activity}}{\text{\# of days either } t \text{ or } v \text{ is active}} \quad (7)$$

$$sr_{tv} = 1 - \left| \max\left(1, \frac{\text{\# of followers of } t}{\text{\# of followings of } t}\right) - \max\left(1, \frac{\text{\# of followers of } v}{\text{\# of followings of } v}\right) \right| \quad (8)$$

$$fr_{tv} = 1 - \left| \frac{\text{\# of fraud tweets of } t}{\text{\# of tweets of } t} - \frac{\text{\# of fraud tweets of } v}{\text{\# of tweets of } v} \right| \quad (9)$$

$$me_{tv} = \frac{\text{common mentions between } t \text{ and } v}{\text{total mentions of } t \text{ and } v} \quad (10)$$

For all the above four features, a higher value represents a closer connection between the pair of nodes. For a source node $u$, we generate a random walk of fixed length $k$. The $i^{th}$ node $c_i$ of a random walk starting at node $c_0$ is generated with the distribution:

$$P(c_i = t \mid c_{i-1} = v) = \begin{cases} \mathcal{B}_{vt}, & if (v, t) \in E \\ 0, & otherwise \end{cases} \quad (11)$$

where $\mathcal{B}_{vt}$ is the normalized transition probability between nodes $v$ and $t$. The transition probability are computed based on the spam dynamics so that the source node has equivalent spam dynamics with its neighborhood nodes.

**Algorithm 1:** ENWalk $(G, d, \lambda, l, k, [p, q, r, s])$

**Input:** graph $G(V, E, W, X)$
embedding dimensions $d$
walks per node $\lambda$
walk length $l$
context size $k$
tweet parameters $p, q, r, s$
**Output:** matrix of latent features $F$

1. $(CT, SR, FR, ME)$ = Preprocess$(G, p, q, r, s)$
2. Initialize $walks$ to empty
3. **for** $i = 1$ to $\lambda$ **do**
4.    **for** each $v_i \; \varepsilon \; V$ **do**
5.       Initialize $walk$ to $v_i$
6.       **for** $j = 1$ to $l$ **do**
7.          $x$ = GetEquivalentNeighbor$(G, CT, SR, FR, ME, walk[j], W)$
8.          Append $x$ to $walk$
9.       Append $walk$ to $walks$
10. $F$ = StochasticGradientDescent$(k, d, walks)$

We define four parameters which guide the random walk. Consider that a random walk just traversed edge $(t, v)$ to now reside at node $v$. The walk now needs to decide on the next step so it evaluates the transition probabilities on edges $(v, x)$ leading from $v$. We set the transition probability to $\mathcal{B}_{vx} = \alpha_{pqrs}(t, v, x) . w_{vx}$, where

$$\alpha_{pqrs}(t, v, x) = p.(ct_{tv} + ct_{vx}) + q.(sr_{tv} + sr_{vx}) + r.(fr_{tv} + fr_{vx}) + s.(me_{tv} + me_{vx}) \quad (12)$$

where the parameters $p, q, r, s$ are used to prioritize the tweet dynamics. To select the next node, the random walk is biased towards the nodes which have similar tweet dynamics to both the current node and the previous node in the random walk.

### 5.4 Algorithm: ENWalk

Algorithm 1 details our entire scheme. We start with $\lambda$ fixed length random walks at each node $l$ times. To obtain each walk, we use *GetEquivalentNeighbor*, the random sampler that samples the node based on the transition probabilities computed in equation 12. It is worth noting that the tweet dynamics between the nodes $(CT, SR, FR, ME)$ defined in equation 7, 8, 9, 10 respectively can be pre-computed. Once, we have random walks we can obtain $d$ dimensional numeric features using the optimization function in equation 6 with a window size of $k$. The three phases preprocessing, random sampling and optimization are asynchronous so that ENWalk is scalable.

## 6 Experiment

We applied ENWalk to twitter dataset to evaluate its effectiveness. In this section, we discuss the baseline methods and compare with ENWalk for classification and ranking.

### 6.1 Baseline Methods

For classification, we compare our model with two graph embedding methods: Deepwalk and node2vec. We use PageRank and Markov Random Field (MRF) approaches for ranking of spam nodes. We did not use feature extraction techniques like [1] as they only use the node features without using the graph structure.

**Table 1. Propagation Matrix for (S)pammer, (M)ixed, (N)on-Spammer**

|   | S | M | N |
|---|---|---|---|
| S | 0.80 | 0.40 | 0.025 |
| M | 0.15 | 0.50 | 0.125 |
| N | 0.05 | 0.10 | 0.850 |

**Deepwalk** [13]. It is the first approach to integrate the language modeling for network feature representation. It generates uniform random walks equivalent to sentences in the language model.

**Node2vec** [13]. It is another representation learning for nodes in the network. It extends the language model of random walks employing a flexible notion of neighborhood. It designs a biased random walk using BFS and DFS neighborhood discovery.

**PageRank Models.** PageRank is a popular ranking algorithm that exploits the link-based structure of a network graph to rank the nodes of the graph.

$$PR = (1 - \alpha) * M * PR + \alpha * p \tag{13}$$

where $M$ is transition probability matrix, $p$ represents the prior probability with which a random surfer surfs to a random page and $\alpha$ is damping factor. For variations of PageRank, we vary the values of $M$ and $p$ using trustworthiness of a user. Trustworthiness ($f_{Trust}$) is using a set of features (# of Blacklist URL, # of tweets, # of mentions, # of duplicate tweets, # of tweets containing adult/bad words, # of tweets containing violent words, # of tweets containing promotional words and the total time of activity for the user). We manually labeled $f_{Trust}$ score of 800 users (400 non-suspended and 400 suspended). We gave a real-valued trustworthiness score between 0 and 1. A value closer to 0 means the user is most likely a spammer. We then obtain the weight of the features by learning linear regression model on the users.

- **Traditional PageRank** We use the default PageRank settings for $M$ and $p$.
- **Trust Induced and Trust Prior:** Transition matrix $M$ is modified as $M_{uv} = \mathrm{M}_{uv} * f_{Trust}(v), \forall u, \forall v$ and $f_{Trust}(v)$ is used as prior probability.

**Markov Random Field Models.** Markov Random Fields are undirected graphs (and can be cyclic) that satisfy the three conditional independence properties (Pairwise, Local, and Global). For the inference, we use the Loopy Belief Propagation algorithm. Inspired by spam detection in [2], we define 3 hidden states {Spammer, Mixed, Non-Spammer} and the Propagation Matrix is used as in Table 1. Logically, spammers follow other spammers more (hence 0.8 probability) and non-spammers tend to follow other non-spammers. We also include the mixed state to include those users who are difficult to categorize spammers or non-spammers.

### 6.2 Node Classification

We obtained the feature representations from three different algorithms: ENWalk, node2vec and DeepWalk using the settings used in node2vec and DeepWalk. All the feature learnings are unsupervised. Similar to node2vec and DeepWalk, we used $d = 128$, $\lambda = 10$, $l = 80$, $k = 10$. We found that the parameters $d, \lambda, l, k$ are sensitive in a similar style to node2vec and DeepWalk. We used each feature representation as an example for standard SVM classifier. We used 10-fold cross-validation using balanced data obtained from sub-sampling of the negative class. From the classification results in Table 2, ENWalk performs better. It has higher precision, recall, F1-score and accuracy due to the biased random walks.

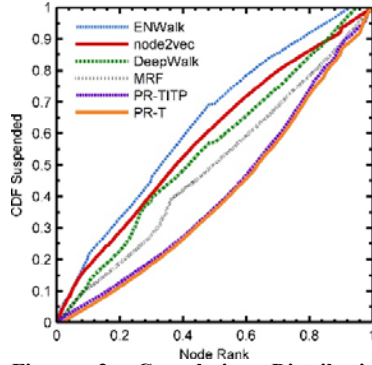

Figure 2. Cumulative Distribution Function of Suspended Nodes

Table 2. Classification Results: Precision (P), Recall (R), F1-score (F) and Accuracy (A)

| Model | P | R | F | A |
|---|---|---|---|---|
| DeepWalk | 0.44 | 0.49 | 0.46 | 0.51 |
| Node2vec | 0.46 | 0.53 | 0.49 | 0.57 |
| ENWalk | 0.59 | 0.66 | 0.62 | 0.71 |

### 6.3 Node Ranking

Table 3. Ranking Results: Area Under CDF Curve (AUC) and Precision@100(P@100)

| Model | AUC | P@100 |
|---|---|---|
| PR-T | 0.4059 | 0.02 |
| PR-TITP | 0.4181 | 0.03 |
| MRF | 0.4944 | 0.02 |
| DeepWalk | 0.5502 | 0.05 |
| node2vec | 0.5836 | 0.05 |
| **ENWalk** | **0.6335** | **0.12** |

We use two metrics to evaluate the ranking results: *Cumulative Distribution Function of Suspended Users* and *Precision@n*. We rank all the nodes in the graph and provide a node rank percentile. For each node rank percentile, we compute the number of suspended users in that percentile. We plot the cumulative distribution function for those suspended users. We also calculate the Area Under Curve (AUC) for the CDF. The higher the area the better the model. Precision@n of Suspended Users evaluates how many top n nodes suggested by a model are actually the suspended users. This is effective to screen the nodes that are probable being spammers.

To evaluate the ranking performance of ENWalk, we use Logistic Regression on the features obtained from the model. We compare our model with PageRank and Markov Random Field models. We present the CDF in Fig 2. We can see that ENWalk outperforms all the baseline models. We also computed the AUC and precision@100 (Table 3). A higher AUC and precision@100 signifies the ability to profile the top spammers.

## 7 Conclusion

We studied the problem of identifying spammers in Twitter who are involved in malicious attacks. This is very much important as it has many practical applications in today's world where almost everyone is actively social online. This paper proposed a method of spam detection in Twitter that makes use of the online network structure and information shared. This data driven approach is important as there is a lot of data of social medias online these days. We demonstrated the helpfulness of biased random walks in learning node embedding that can be used for classification and ranking tasks.

**Acknowledgements**: This work is supported in part by NSF 1527364. We also thank anonymous reviewers for their helpful feedbacks.